\documentclass[lettersize,journal]{IEEEtran}
\usepackage{amsmath,amsfonts}
\usepackage{algorithm}
\usepackage{algorithmicx}
\usepackage[noend]{algpseudocode}
\usepackage{array}
\usepackage{textcomp}
\usepackage{stfloats}
\usepackage{url}
\usepackage{verbatim}
\usepackage{graphicx}
\usepackage{cite}
\usepackage[inline]{enumitem} 

\usepackage{nth}
\usepackage{bm}
\usepackage{multicol}
\usepackage{multirow}
\usepackage{booktabs}

\usepackage{graphicx}
\usepackage{subcaption}
\usepackage{caption}
\usepackage{tikz}
\usetikzlibrary{arrows.meta, positioning, calc, plotmarks}

\usepackage{xcolor}

\definecolor{classA}{RGB}{0,114,178}   
\definecolor{classB}{RGB}{213,94,0}    
\definecolor{classC}{RGB}{0,158,115}   

\definecolor{unknown}{RGB}{160,160,160} 
\definecolor{boundary}{RGB}{200,200,200}

\definecolor{tab0}{HTML}{1f77b4}
\definecolor{tab1}{HTML}{ff7f0e}
\definecolor{tab2}{HTML}{2ca02c}
\definecolor{tab3}{HTML}{d62728}
\definecolor{tab4}{HTML}{9467bd}
\definecolor{tab5}{HTML}{8c564b}
\definecolor{tab6}{HTML}{e377c2}
\definecolor{tab7}{HTML}{7f7f7f}
\definecolor{tab8}{HTML}{bcbd22}
\definecolor{tab9}{HTML}{17becf}

\newcommand{\swatch}[1]{\tikz[baseline=-0.5ex]{\fill[#1] (0,0) rectangle (0.25,0.25);}}

\begin{document}


\title{Learning Kernels by Alignment for Multiclass Bayes Classification}

\author{Hollan~Haule and~Javier~Escudero,~\IEEEmembership{Senior~Member,~IEEE}

\thanks{
HH acknowledges an EPSRC postdoctoral pathways for growth in the UK grant (part of UKRI3036). JE acknowledges a Cure Syngap1 -- Syngap Research Fund grant and an EPSRC grant (UKRI1659).
}

\thanks{Manuscript received April 19, 2021; revised August 16, 2021.}}

\markboth{Journal of \LaTeX\ Class Files,~Vol.~14, No.~8, August~2021}%
{Shell \MakeLowercase{\textit{et al.}}: A Sample Article Using IEEEtran.cls for IEEE Journals}


\maketitle

\begin{abstract}
Kernel methods separate data representation from decision-making, but typically require the kernel to be chosen in advance. We show that this kernel can instead be learned by alignment, and develop the resulting framework through the recently introduced Collaborative Learning and Inference (CLaI). We show that Collaborative Learning can be viewed as a kernel alignment process, in which an embedding is trained so that its induced similarity matches a label-derived target kernel. We also prove that Collaborative Inference is equivalent to kernel Bayes classification with Parzen-window density estimation. Motivated by these perspectives, we generalise CLaI by replacing cosine similarity with a learned Mahalanobis distance and extend it to multiclass classification. On CIFAR-10, PathMNIST, and SleepEDF, the Mahalanobis formulation improves accuracy, converges faster, and yields lower calibration error than the cosine-based variant. Auxiliary experiments further support these connections, showing that CLaI produces latent signals of the same form as a Gaussian process, while achieving competitive calibration on sepsis prediction. Together, these results establish a principled learned-kernel framework that unifies representation learning, kernel alignment, and Bayesian classification, and extends naturally to the multiclass setting.

\end{abstract}


\section{Introduction}\label{sec:intro}

Kernel methods are a popular machine learning paradigm due to their ability to implicitly map input data to a high-dimensional feature space, known as a Reproducing Kernel Hilbert Space (RKHS). In this feature space, traditional kernel-based algorithms can efficiently discover patterns in the data. Essentially, kernel methods perform two distinct tasks: 
\begin{enumerate*}[label=(\roman*)]
\item mapping data to a feature space, and
\item performing pattern discovery or learning in that space \cite{wang_overview_2015, shawe-taylor_kernel_2004}.
\end{enumerate*}


Despite their strengths, kernel methods face significant challenges. One major issue is the selection and tuning of the kernel function, which is often a non-trivial task that demands extensive user intervention. For example, support vector machines (SVMs) do not only require solving a quadratic program to determine the decision function, but also rely on cross-validation to tune the kernel parameters \cite{du_exploring_2024,wilson_deep_2016}. This coupling of hyper-parameter tuning with the learning process makes model selection computationally expensive and complex \cite{wang_overview_2015}.


Deep kernel learning (DKL) methods leverage deep learning to automatically learn the appropriate kernel function and its hyper-parameters directly from the data. For example, in \cite{le_deep_2016} a deep network is designed to learn a symmetric kernel function, with training performed via a logistic regression layer in a semi-binary classification task to capture pairwise similarity. Similarly, \cite{Wilson2016DeepLearning} embeds the deep learning module within a base kernel framework, enabling joint learning of both the deep feature representation and the kernel hyper-parameters using the Gaussian process marginal likelihood. Compartmentalising the deep learning function within the base kernel also facilitates learning a rich, infinite basis function representation of the data through the kernel \cite{Wilson2016DeepLearning}.


In parallel, kernel evaluation metrics such as kernel alignment have been introduced \cite{cristianini_kernel-target_2001, wang_overview_2015, insulla_towards_2024} to aid in the model selection process. Kernel alignment measures the similarity between the kernel matrix and a target similarity matrix. Moreover, because kernel alignment is independent of the underlying learning machine, it is simple and efficient to compute \cite{wang_overview_2015}. In addition, kernel alignment exhibits a strong concentration property \cite{cristianini_kernel-target_2001, wang_overview_2015}, with empirical estimates maintaining low variance across different data subsets. This stability suggests that optimising kernel alignment, even with mini-batches, can provide reliable gradient estimates and consequently improve the training stability of DKL models.



While many kernel-based methods have been successful in binary classification, extending these techniques to multiclass problems remains a significant research interest \cite{pinheiro_quantum_2026,hanriot_multiclass_2025}. Multiclass classification techniques generally fall into one of two categories: those that employ a single predictor function to manage all classes and those that decompose the problem into multiple binary classification tasks (using strategies such as one-vs-one or one-vs-rest) \cite{Rifkin2004InClassification, wang_overview_2015, crammer_algorithmic_2001, hsu_comparison_2002}.



In this manuscript, we build on Collaborative Learning and Inference  (\textbf{CLaI})\cite{Haule2025CLaI:Prediction}, showing it is a faithful instantiation of kernel methods that cleanly separates data mapping from decision-making. In our framework, \textbf{collaborative learning}, which we show is equivalent to a kernel alignment process, directly learns a mapping function $\phi$ \textit{and} the associated kernel hyper-parameters from the data. \textbf{Collaborative inference} then determines the predictor in the latent space through a similarity-weighted combination of reference labels, a solution we show to be equivalent to a kernel Bayes classifier with Parzen-window density estimation.

Moreover, we extend CLaI to the multiclass setting by conditioning $\phi$ on a target class under a one-versus-rest strategy, despite the conventional restriction of kernel alignment to binary tasks \cite{wang_overview_2015}. To further improve multiclass performance, we introduce a Mahalanobis-distance base kernel as the similarity function, inspired by DKL \cite{wilson_deep_2016}. We further position Collaborative Learning within the family of learned-kernel methods, where the similarity function is learned directly from data through kernel alignment rather than selected a priori.

The main contributions of this work include:
\begin{itemize}
        \item Proving the theoretical connection between collaborative inference and the Bayes classifier.
        \item Demonstrating that collaborative learning is essentially optimising a kernel alignment during training.
        \item Extending collaborative learning to use a generalised distance function based on the Mahalanobis distance.
        \item Presenting an extension of the collaborative learning framework (CLaI) for multiclass problems, validated on CIFAR‑10, PathMNIST, and SleepEDF.
    \end{itemize}

This manuscript is organised as follows: Section~\ref{sec:background} covers the background, Section~\ref{sec:theoretical_analysis} covers a formal description of CLaI, and shows its connection to kernel alignment, Bayes classifier and extension to multiclass. Then Section~\ref{sec:experiments}, \ref{sec:discussion}, \ref{sec:limitations}, and \ref{sec:conclusion}, respectively, cover experiments and results, discussion, limitations, and conclusions.

\section{Background on Related Work}\label{sec:background}
\subsection{Kernel Methods fundamentals}
Many supervised Kernel methods can be formulated as regularised empirical risk minimisation problems in Reproducing Kernel Hilbert Spaces (RKHS), \eqref{eq:regularisedER} as 
\begin{equation}\label{eq:regularisedER}
    \min_{f\in\mathcal{H}} \; \sum_{i=1}^n L(y_i, f(\textbf{x}_i)) + \lambda \|f\|_{\mathcal{H}}^2
\end{equation}
where the Representer theorem shows that the minimiser $\hat{f}$ takes the form 
\begin{equation}\label{eqn:representer_theorem}
    \hat{f}(\textbf{x}) = \sum_{i=1}^n \alpha_i \, k(\textbf{x}, \textbf{x}_i),
\end{equation}
a finite linear combination of the kernel functions, $k(\textbf{x}_i,\cdot)$ centred at the training points, $\textbf{x}_i$. In \eqref{eq:regularisedER},  $\mathcal{H}$ is the RKHS induced by the kernel $k$, \(L\) is a loss function, \(\lambda > 0\) is a regularisation parameter, and \(\{(\textbf{x}_i, y_i)\}_{i=1}^n\) is the training set. The regularised empirical loss splits the risk into two parts: the Kernel, $\|f\|_{\mathcal{H}}^2$ and the task $L(y_i,\cdot)$. Moreover, based on Aronszajn's theorem \cite{Murphy2022ProbabilisticIntroduction}, the kernel induced RKHS is such that
\begin{equation}\label{eqn:aronzjan}
    k(\textbf{x},\textbf{x}') = \langle \phi(\textbf{x}), \phi(\textbf{x}') \rangle_{\mathcal{H}} = \langle k(\cdot,\textbf{x}), k(\cdot,\textbf{x}') \rangle_{\mathcal{H}}
\end{equation}
where $\phi(\textbf{x})$ is a mapping to the Hilbert space unique to the kernel.

\subsection{Deep Kernel Learning (DKL)}
As highlighted in Section~\ref{sec:intro}, a number of DKL exist in literature \cite{nikhitha_deep_2021,cho_kernel_2009, wang_bridging_2021}. The most relevant for our work is the work by Wilson \MakeLowercase{\textit{et al.}} \cite{wilson_deep_2016}, which embeds a deep learning module, $\phi_{\theta}(\textbf{x})$, within a base kernel
\begin{equation}\label{eq:dkl_wilson}
    k_{\theta}(\textbf{x},\textbf{x}') \rightarrow k_{\text{base}}(\phi_{\theta}(\textbf{x}),\phi_{\theta}(\textbf{x}')).
\end{equation}

This approach allows learning a unique mapping to the Hilbert space and hyperparameters directly from the data. Moreover, Mercer's theorem implies that encapsulating the mapping $\phi$ inside the base kernel leads to a model that combines learned representations with an infinite-dimensional function space, trained jointly via marginal likelihood. 


\subsection{Kernel Alignment}
Kernel Alignment \cite{cristianini_kernel-target_2001} measures the similarity between two kernel matrices, as an evaluation metric for kernel model selection \cite{wang_overview_2015}. Consider a binary classification task and training dataset $D = \left\{ (\textbf{x}_i,y_i)\right\}_{i=1}^{n}$ and $y_i \in \left\{ \pm 1\right\}$. Given a kernel matrix $K$ and a target $T = \textbf{y}\textbf{y}^T$, where $\textbf{y} = \left ( y_1,\cdots, y_n  \right )^T$, Kernel alignment is computed as
\begin{equation}
    A(K,T) = \frac{\left< K,\textbf{y}\textbf{y}^T\right>_F}{\sqrt{\left< K,K\right>_F \left<\textbf{y}\textbf{y}^T,\textbf{y}\textbf{y}^T\right>_F }} =
\end{equation}
\[
= \frac{\textbf{y}^TK\textbf{y}}{n\left\| K\right\|_F}
\]
where $\left< .,.\right>_F$ and $\left\| .\right\|_F$ are the Frobenius inner product and norm.

Maximising Kernel alignment is achieved by minimising the distance $d(K,T)$ of a Kernel $K$ to the target kernel $T$
\begin{equation}\label{eq:kernel_alignment_distance}
    d(K,T) = \left\| \frac{K}{\left\| K\right\|}_F - \frac{\textbf{y}\textbf{y}^T}{\left\| \textbf{y}\textbf{y}^T\right\|}_F \right\|= \sqrt{2 -2A(K,T)}.
\end{equation}

\subsection{Bayes Classifier}\label{sec:bayes_classifier}

In statistical decision theory \cite{Nordhausen2009TheFriedman}, the objective is to construct a predictor that minimises the expected misclassification error. Under the assumption that the class-conditional distributions are known (or can be accurately estimated), the Bayes classifier provides the optimal decision rule for binary classification:

\begin{equation}\label{eq:bayes_classifier}
y = \operatorname{sgn}\left( f_{+}(\mathbf{x})P_{+} - f_{-}(\mathbf{x})P_{-} \right).
\end{equation}

Here, $f_{+}(\mathbf{x})$ and $f_{-}(\mathbf{x})$ denote the class-conditional probability density functions (likelihoods) for the positive and negative classes, respectively, while $P_{+}$ and $P_{-}$ denote the corresponding prior probabilities, where $f_{\pm}(\mathbf{x}) = f(\mathbf{x}\mid C=\pm1)$ and $P_{\pm} = \Pr(C=\pm1)$.







\section{Approach and Theoretical Insights}\label{sec:theoretical_analysis}
In this section, we split the contributions on this work in to three main parts:
\begin{enumerate*}[label=(\roman*)]
\item Describing and establishing the connection between CLaI and kernel methods and the Bayes classifier;
\item Introducing the Mahalanobis distance kernel as part of DKL with collaborative learning; and 
\item Extending CLaI to multiclass classification.
\end{enumerate*}

\subsection{CLaI: Collaborative Learning and Inference}\label{sec:clai}
In our previous work \cite{Haule2025CLaI:Prediction}, we introduced CLaI as a two-step binary classification-based approach originally designed for time series. The assumption behind CLaI \cite{Haule2025CLaI:Prediction} is that there is a latent space where embeddings of samples belonging to the same class are clustered while different classes are pushed apart. The two steps involved are: 
\begin{enumerate*}[label=(\roman*)]
\item \textit{Collaborative Learning}: learning a mapping function $\phi$ to an embedding space, where similar samples are clustered; and
\item \textit{Collaborative Inference}: a new sample is assigned a class based on its proximity to a reference set of samples in the embedding space
\end{enumerate*}. Here, a more general formulation is presented, suitable for other types of data.

\subsubsection{Collaborative Learning}\label{sec:collaborative_learning}
Collaborative Learning is based on the underlying assumption that there exists a latent space such that similar samples are clustered together. To setup the problem, we are given a dataset $\mathcal{D} =  \left \{ \left ( \textbf{x}_i,y_i \right ) \right \}_{i=1}^{N}, \quad \textbf{x}_i \in \mathbb{R}^D,\; y_i \in \{-1, +1\}$, partitioned into $M = \lfloor |\mathcal{D}| / B \rfloor$ mini-batches, $\{(X_t,\textbf{y}_t)\}_{t=1}^M$, where $X_t = (\textbf{x}_1,\cdots, \textbf{x}_B)^T \in \mathbb{R}^{B \times D}, \textbf{y} = (y_1,\cdots,y_B)^T \in \left \{ -1,+1 \right \}^B$, and batch size $B$.

We define a parameterised embedding model
\[
\phi_{\theta}: \mathbb{R}^D\to \mathbb{R}^L,
\]
which maps each batch $X_t$ to latent representations
\[
Z_t = \phi_{\theta}(X_t) \in \mathbb{R}^{B \times L},
\]
where $L$ is the embedding dimension and $\theta$ denotes the model parameters.


Let $t$ and $r$ denote two (distinct) mini-batches, referred to as the anchor and reference batches, respectively. We define a pairwise similarity function
\[
k_{\theta} : \mathbb{R}^D \times \mathbb{R}^D \to [-1,+1].
\]




$k_{\theta}$ induces the cross-batch similarity matrix
\[
K_{\theta}(X_t, X_r) \in \mathbb{R}^{B \times B},
\]
whose $(p,q)$-$th$ entry is
\[
(K_{\theta}(X_t, X_r))_{pq} = k_{\theta}((X_t)_{p,:},(X_r)_{q,:}).
\]

In our previous implementation of CLaI \cite{Haule2025CLaI:Prediction}, $k_{\theta}$ is chosen as the cosine similarity between latent embeddings. Consequently,

\begin{equation}\label{eq:cosine_similarity}
    S = K_{\theta}(X_t, X_r) = \frac{Z_t Z_r^T}
{\|Z_t\| \, \|Z_r\|},
\end{equation}
where the normalization is applied row-wise.

The desired similarity structure between the two batches is encoded by the target matrix
\[
T = \mathbf{y}_t \mathbf{y}_r^\top
\]
where
\[
T_{ij} =
\begin{cases}
+1, & \text{if } y_i = y_j,\\[0.2mm]
-1, & \text{if } y_i \neq y_j.
\end{cases}
\]

Thus, samples from the same class are encouraged to have high similarity, while samples from different classes are encouraged to have low similarity.

Collaborative Learning trains the embedding model by aligning the predicted similarity matrix $S$ with the target matrix $T$. Specifically, the parameters $\theta$ are learned by minimising

\begin{equation}\label{eq:clai_loss}
    \ell_{\theta} = \frac{1}{B^2} \| S - T \|_F^2.
\end{equation}


During each optimisation step, an anchor batch $t$ is compared with each reference batch $r \neq t$. The resulting losses are accumulated before updating the model parameters. The complete training procedure is summarised in Algorithm~\ref{algorithm_collaborative_learning}. Duplicate batch pairs $(t,r)$ and $(r,t)$, as well as self-comparisons $(t,t)$, are omitted.


The loss is a matrix alignment objective since
\[
\| S - T \|_F^2 = \langle S - T, S - T\rangle_F = \left\| S\right\|_F^2 + \left\| T\right\|_F^2 - 2\langle S, T \rangle_F.
\]

When $S$ and $T$ are normalised, this reduces to
\[
d(S,T)^2 = 2 - 2\langle S, T \rangle_F,
\]
which is precisely the kernel alignment distance defined in \eqref{eq:kernel_alignment_distance}. Hence, Collaborative Learning can be interpreted as directly optimising kernel alignment between the learned similarity matrix and the label-induced target matrix.

\begin{algorithm}
\caption{Collaborative Learning -- Binary}
\begin{algorithmic}[1]

\Require Dataset $\mathcal{D} = \{(x_i, y_i)\}_{i=1}^N$, batch size $B$, epochs $T$, Learning rate $\eta$
\State Initialization, model = $\phi_{\theta}(\cdot)$
\For{$\tau = 1$ to $T$}
    \State Shuffle $\mathcal{D}$
    \State Partition $\mathcal{D}$ into $M = \lfloor |\mathcal{D}| / B \rfloor$ batches $\{(X_t,\textbf{y}_t)\}_{t=1}^M$
    \For{$t = 1$ to $M-1$}
    \State $\nabla_\theta \gets \textit{zero\_grad}(\theta)$
        \State $Z_t = \phi_{\theta}(X_t)$
        \State Form label vector $\mathbf{y}_t$
        
        \For{$r = t+1$ to $M$}

            
            \State $Z_r = \phi_{\theta}(X_r)$
            
            \State Form label vector $\mathbf{y}_r$
            
            \State Compute $S= \frac{Z_t Z_r^T}
            {\|Z_t\| \, \|Z_r\|}$
            
            \State Compute target matrix $T = \mathbf{y}_t \mathbf{y}_r^\top$
            
            \State $\ell \gets  \frac{1}{B^2} \|S - T\|_F^2$
            \State $\nabla_\theta \gets \textit{backward}(\ell,\ \text{retain\_graph}=\text{True})$
            
        \EndFor
        \State $\theta \gets \theta - \eta \nabla_\theta$
    \EndFor
\EndFor

\end{algorithmic}
\label{algorithm_collaborative_learning}
\end{algorithm}

\subsubsection{Collaborative Inference}
Collaborative Inference uses the embedding model learned by Algorithm~\ref{algorithm_collaborative_learning} to predict labels for unseen samples. Let
\[
\mathcal{R} =  \left \{ \left ( \textbf{x}_r,y_r \right ) \right \}_{r=1}^{M}, \quad \textbf{x}_r \in \mathbb{R}^D,\; y_r \in \{-1, +1\},
\]
denote a labelled reference set, and
\[
\mathcal{T} =  \left \{ \textbf{x}_p \right \}_{p=1}^{P}, \quad \textbf{x}_p \in \mathbb{R}^D
\]
denote a set of unlabelled test samples. The reference set may simply be the same dataset $\mathcal{D}$ used during Collaborative Learning. Given the trained embedding model $\phi_{\theta}(\cdot)$, the objective is to predict the test labels

\[
\textbf{f} = \left ( f_1,\cdots, f_P  \right )^T.
\]

During training, the embedding model is optimised so that cross-batch similarity matrices approximate the corresponding label agreement matrices. Consequently, for a batch of test samples and a batch of reference samples, we expect the learned similarity matrix to satisfy
\[
S \approx  \textbf{f}\textbf{y}^T,
\]
leading to the predictor
\begin{equation}\label{eq:collaborative_inference}
\mathbf f = S\mathbf y
(\mathbf y^\top\mathbf y)^{-1},
\end{equation}
where $\textbf{y}$ contains the known labels of the reference samples and \textbf{f} denotes the unknown labels of the test samples. The predicted class labels are then obtained by thresholding
\[
\hat{\textbf{y}} = \operatorname{sgn}(\textbf{f}).
\]

The complete inference procedure is summarised in Algorithm~\ref{inference_algorithm}.





\begin{algorithm}
\caption{Collaborative inference -- Binary}
\begin{algorithmic}[1]
\State \textbf{Input:} $\mathcal{R} = \{(x_i, y_i)\}_{i=1}^R$, $\mathcal{T} =  \left \{ \textbf{x}_p \right \}_{p=1}^{P}$, batch size $B$
\State Load model = $\phi_{\theta}$
\State $\mathbf{y} \gets \begin{bmatrix} \text{ } \end{bmatrix}, S \gets \begin{bmatrix} \text{ } \end{bmatrix}$
\State Partition $\mathcal{R}$ into $R = \lfloor |\mathcal{D}| / B \rfloor$ batches $\{(X_r,\textbf{y}_r)\}_{r=1}^R$
Form $X = (\textbf{x}_1,\cdots, \textbf{x}_P)^T$ from $\mathcal{T}$
\State $Z = \phi(X)$
\For{$r = 1$ to $R$}
    \State $Z_r = \phi(X_r)$
    \State Compute $\grave{S} = \frac{Z Z_r^T}
            {\|Z\| \, \|Z_r\|}$
    \State Form label vector $\grave{\mathbf{y}}$
    \State $S \gets \begin{bmatrix} S & \grave{S} \end{bmatrix}$
    \State $\mathbf{y} \gets \begin{bmatrix} \mathbf{y} \\ \mathbf{y}_r \end{bmatrix}$
\EndFor
\State $\textbf{f} = S\textbf{y}\left (  \textbf{y}^{T}\textbf{y}\right )^{-1}$
\State $\textbf{return}$  $\operatorname{sgn}(\textbf{f})$
\end{algorithmic}
\label{inference_algorithm}
\end{algorithm}

\subsection{Collaborative Inference and Bayes Classifier}\label{sec:collaborative_inference_bayes}
We now show that the Collaborative Inference decision rule can be interpreted as a kernel approximation to the Bayes classifier introduced in Section~\ref{sec:bayes_classifier}. In particular, we demonstrate that the prediction rule in \eqref{eq:collaborative_inference} has the same form as a Bayes classifier when the class-conditional densities are estimated using kernel (Parzen window) density estimation.

Consider predicting the label of a single test sample $\textbf{x}$ using a labelled reference set
\[
\mathcal{R} =  \left \{ \left ( \textbf{x}_n,y_n \right ) \right \}_{n=1}^{N}.
\]

In this case, the similarity matrix in \eqref{eq:collaborative_inference} reduces to the similarity vector
\begin{equation}
S_n = K(\mathbf x,\mathbf x_n) = \langle
\phi(\mathbf x),
\phi(\mathbf x_n)
\rangle.
\end{equation}

\begin{align}
y
&=
\operatorname{sgn}
\left(
\frac{\sum_n K(\mathbf x,\mathbf x_n)y_n}
{\sum_n y_n^2}
\right)
\\
&=
\operatorname{sgn}
\left(
\frac1N
\sum_n
K(\mathbf x,\mathbf x_n)y_n
\right)
\\
&=
\operatorname{sgn}
\left(
\frac{N_+}{N}
\sum_{y_n=+1}
K(\mathbf x,\mathbf x_n)
-
\frac{N_-}{N}
\sum_{y_n=-1}
K(\mathbf x,\mathbf x_n)
\right)\label{eqn:kernel_bayes}.
\end{align}

Comparing \eqref{eqn:kernel_bayes} with the Bayes classifier in \eqref{eq:bayes_classifier} reveals a direct correspondence. The empirical class frequencies
\[
\frac{N_+}{N}, \quad \frac{N_-}{N}
\]
estimate the prior probabilities $P_{+}$ and $P_{-}$	, while the kernel summations
\[
\sum_{y_n=\pm1}
K(\mathbf x,\mathbf x_n)
\]
are Parzen window estimates of the class-conditional densities $f_{\pm}(\mathbf{x})$. Consequently, Collaborative Inference implements a kernel Bayes classifier in the learned embedding space.

This establishes Collaborative Inference as a kernel Bayes classifier operating on a learned similarity function. Unlike conventional Parzen classifiers, which employ a predefined kernel, the similarity kernel is induced by the embedding model learned during Collaborative Learning through the kernel alignment objective.
\subsection{DKL with Collaborative Learning}\label{sec:mahalanobis_kernel}
Following the Deep Kernel Learning formulation of Wilson $\textit{et al}$. \cite{wilson_deep_2016}, we modify collaborative learning to use Mahalanobis distance \cite{Weinberger2009DistanceClassification}, inside an exponential base kernel, instead of cosine similarity. We then rescale it to have values between $-1$ and $+1$ so that it is compatible with the target matrix:
\begin{equation}\label{eqn:mahalanobis}
    k_{\theta}\left ( \textbf{x}_i, \textbf{x}_j \right ) =2 \exp\left(-(\textbf{z}_i-\textbf{z}_j)^TM(\textbf{z}_i-\textbf{z}_j)\right) - 1
\end{equation}
where $\textbf{z} = \phi(\textbf{x})$ $M$ is positive semidefinite, called a metric matrix, and $L$ is a lower triangular matrix of real values according to Cholesky decomposition, defined as $M = L^TL$. We learn both $\phi_{\theta}$ and $M_{\theta}$ directly from data using neural networks.

We compare the performance of this kernel with cosine similarity using CLaI for multiclass prediction in Section~\ref{sec:ablation_similarity_function}.

\subsection{Extending CLaI to Multiclass classification}
CLaI as presented in section~\ref{sec:clai} works only for binary class problems. Here we extend CLaI to work in multiclass setting. As described in \cite{wang_overview_2015}, one common approach involves decomposing a multiclass task into several binary classification tasks. Our approach is based on the one-versus-rest (1-v-r) strategy, also known as one-vs-all in other literature \cite{Rifkin2004InClassification}. 

\subsubsection{Collaborative Learning}
\begin{figure*}
  \centering
  \resizebox{0.7\textwidth}{!}{\begin{tikzpicture}[
    >=Stealth,
    every node/.style={font=\small},
]

\def\r{1.6} 
\def\gap{5} 

\coordinate (c0) at (0,0);
\coordinate (c1) at (\gap,0);
\coordinate (c2) at (2*\gap,0);

\foreach \c in {c0,c1,c2} {
    \draw[boundary] (\c) circle (\r);
}

\foreach \a in {0,15,...,345} {
    \pgfmathsetmacro{\rad}{0.2 + rnd*1.2}
    \draw[unknown] plot[mark=*, mark size=1.2pt]
        coordinates {($(c0)+(\a:\rad)$)};
}

\foreach \a in {120,135,...,240} {
    \pgfmathsetmacro{\rad}{0.3 + rnd}
    \draw[classA] plot[mark=+, mark size=2.2pt]
        coordinates {($(c1)+(\a:\rad)$)};
}

\foreach \a in {-60,-45,...,60} {
    \pgfmathsetmacro{\rad}{0.3 + rnd}
    \draw[unknown] plot[mark=*, mark size=1.2pt]
        coordinates {($(c1)+(\a:\rad)$)};
}

\node[below=1.7cm of c1] {$k=1$};

\foreach \a in {140,155,...,220} {
    \pgfmathsetmacro{\rad}{0.3 + rnd}
    \draw[classA] plot[mark=+, mark size=2.2pt]
        coordinates {($(c2)+(\a:\rad)$)};
}

\foreach \a in {20,35,...,100} {
    \pgfmathsetmacro{\rad}{0.3 + rnd}
    \draw[classB] plot[mark=x, mark size=2.2pt]
    coordinates {($(c2)+(\a:\rad)$)};
}

\foreach \a in {-100,-85,...,-20} {
    \pgfmathsetmacro{\rad}{0.3 + rnd}
    \draw[unknown] plot[mark=*, mark size=1.2pt]
    coordinates {($(c2)+(\a:\rad)$)};
}

\node[below=1.7cm of c2] {$k=2$};

\draw[->, thick]
    ($(c0)+(0:\r)$) -- 
    node[above] {$\phi(\cdot, k=1)$}
    ($(c1)+(180:\r)$);

\draw[->, thick]
    ($(c1)+(0:\r)$) -- 
    node[above] {$\phi(\cdot, k=2)$}
    ($(c2)+(180:\r)$);

\draw[dotted, thick]
    ($(c2)+(2.2,0)$) -- ++(0.6,0);

\draw[->, thick]
    ($(c2)+(2.8,0)$) -- ++(1.2,0)
    node[midway, above] {$\phi(\cdot, k=C)$};


\matrix[
    above=2cm of c1,
    column sep=1.5cm,
    nodes={anchor=west}
] {
    \node{
        \tikz[baseline=-0.6ex] 
        \draw[unknown] plot[mark=*, mark size=1.5pt] coordinates {(0,0)};
        \hspace{0.4em} Unknown
    }; &

    \node{
        \tikz[baseline=-0.6ex] 
        \draw[classA] plot[mark=+, mark size=2.5pt] coordinates {(0,0)};
        \hspace{0.4em} Class 1
    }; &

    \node{
        \tikz[baseline=-0.6ex] 
        \draw[classB] plot[mark=x, mark size=2.5pt] coordinates {(0,0)};
        \hspace{0.4em} Class 2
    }; \\
};

\end{tikzpicture}}
  
  \caption{Multiclass classification in stages, conditioning the embedding model $\phi(\cdot,k)$ with different classes.}
  \label{fig:clai_multiclass}
\end{figure*}
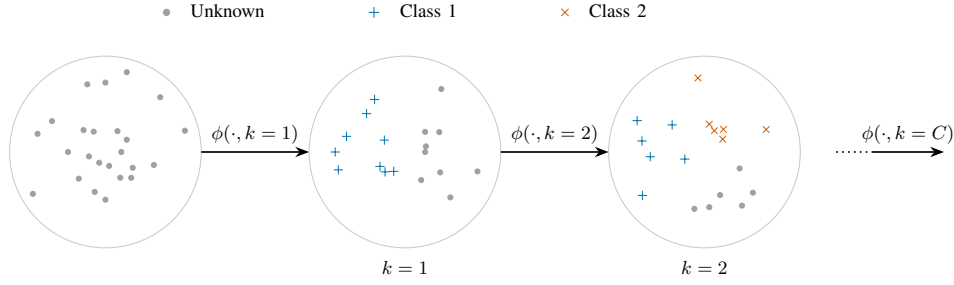

Given a dataset $\mathcal{D} =  \left \{ \left ( \textbf{x}_i,y_i \right ) \right \}_{i=1}^{N}$, $\textbf{x}_i \in \mathbb{R}^D,\; y_i \in \{1,2, \cdots, C\}$ , partitioned into $M = \lfloor |\mathcal{D}| / B \rfloor$ mini-batches, $\{(X_t,\textbf{y}_t)\}_{t=1}^M$, where $X_t = (\textbf{x}_1,\cdots, \textbf{x}_B)^T \in \mathbb{R}^{M \times D}, \textbf{y} = (y_1,\cdots,y_B)^T \in \left \{ -1,+1 \right \}^B$, and batch size $B$. To achieve a 1-v-r strategy with CLaI, we introduce a class conditioning variable, $k$, in the embedding function $Z = \phi_{\theta}(X,k)$. This creates the appearance of having different mapping functions for each class, effectively transforming a multiclass problem into multiple binary tasks, as shown in Fig.~\ref{fig:clai_multiclass}.

Given batch indices $r,t$, we compute the similarity matrix, $S^{(k)} = K_{(\theta)}(Z_t,Z_r) \in \left ( -1,+1\right )^{B \times B}$. We also compute the target matrix $T^{(k)} = \mathbf{y}_t \mathbf{y}_r^\top \in \left \{ -1, +1 \right \}^{B \times B}$ where

\noindent
\begin{minipage}{0.48\columnwidth}
\[
T_{ij}^{(k)} =
\begin{cases}
+1, & \text{if } y_i^k = y_j^k = +1,\\[0.2mm]
0,  & \text{if } y_i^k = y_j^k = -1,\\[0.2mm]
-1, & \text{if } y_i^k \neq y_j^k;
\end{cases}
\]
\end{minipage}
\hfill
\begin{minipage}{0.48\columnwidth}
\[
y_i^k =
\begin{cases}
+1, & \text{if } y_i = k,\\[0.2mm]
-1, & \text{otherwise}.
\end{cases}
\]
\end{minipage}

Note here that to avoid classes that have already been separated being re-clustered, samples belonging to the same negative class are assigned a zero, i.e. essentially treated as ``\textit{unknown}'' and reserved for the next class conditioning.  Full training procedure is listed in Algorithm~\ref{algorithm_collaborative_learning_multiclass}.

{\scriptsize
\begin{algorithm}
\caption{Collaborative Learning -- Multiclass}
\begin{algorithmic}[1]

\Require Dataset $\mathcal{D} = \{(x_i, y_i)\}_{i=1}^N$, batch size $B$, epochs $T$, Learning rate $\eta$, class count $C$
\State Initialization, model = $\phi_{\theta}(\cdot, \cdot)$; $M_{\theta}$
\For{$\tau = 1$ to $T$}
    \For{$k = 1$ to $C$}
        \State Shuffle $\mathcal{D}$
        \State Re-assign labels $\acute{\mathcal{D}} = \{(x_i, y_i^k)\}_{i=1}^N$
        \State Partition $\acute{\mathcal{D}}$, such that $M = \lfloor |\acute{\mathcal{D}}| / B \rfloor$ $\{(X_i,\textbf{y}_i)\}_{i=1}^M$
        \For{$t = 1$ to $M-1$}
        \State $\nabla_\theta \gets \textit{zero\_grad}(\theta)$
            \State $Z_t = \phi_{\theta}(X_t,k)$
            
            \For{$r = t+1$ to $M$}
    
                
                \State $Z_r = \phi_{\theta}(X_r,k)$
                
                
                \State Compute $S^{(k)} = K_{(\theta;M)}(Z_t,Z_r)$
                
                \State Compute target matrix $T^{(k)} = \mathbf{y}_t \mathbf{y}_r^\top$
                
                \State $\ell \gets  \frac{1}{B^2} \|S^{(k)} - T^{(k)}\|_F^2$
                \State $\nabla_\theta \gets \textit{backward}(\ell,\ \text{retain\_graph}=\text{True})$
                
            \EndFor
            \State $\theta \gets \theta - \eta \nabla_\theta$
        \EndFor
    \EndFor
\EndFor

\end{algorithmic}
\label{algorithm_collaborative_learning_multiclass}
\end{algorithm}
}

\subsubsection{Collaborative Inference}
Given a dataset $\mathcal{R} =  \left \{ \left ( \textbf{x}_i,y_i \right ) \right \}_{i=1}^{N}$, $\textbf{x}_i \in \mathbb{R}^D,\; y_i \in \{1,2, \cdots, C\}$ , partitioned into $M = \lfloor |\mathcal{R}| / B \rfloor$ mini-batches, and a set of test samples $\mathcal{T} =  \left \{ \textbf{x}_p \right \}_{p=1}^{P}, \quad \textbf{x}_p \in \mathbb{R}^D$, arranged as $X = (\textbf{x}_1,\cdots, \textbf{x}_P)^T$. Collaborative inference predicts the test labels $\textbf{f} = \left ( f_1,\cdots, f_P  \right )^T$, leveraging cross-similarity as
\[
S^{(k)} \approx \hat{\textbf{y}}\textbf{y}^T \rightarrow S^{(k)} =  \textbf{f}_k\textbf{y}^T \rightarrow \textbf{f}_k = S^{(k)}\textbf{y}\left (  \textbf{y}^T\textbf{y}\right )^{-1}
\]
where $ \hat{\textbf{y}}$ and $\textbf{y}$ are the ideal target and reference label vectors. Following the 1-v-r approach, we compute $\left \{ \textbf{f}_k \right \}_{k=1}^{C}$, before applying softmax on $F = [\textbf{f}_1,\cdots,\textbf{f}_C]$ and selecting the class with the most confident prediction:
\[
F_{i,:} = \operatorname{softmax}(F_{i,:}) \rightarrow \textbf{y}_{pred} = \arg\max_{k} F_{:,k}.
\]
The full collaborative inference algorithm is shown in Algorithm~\ref{inference_algorithm_multiclass}.


{\scriptsize
\begin{algorithm}
\caption{Collaborative inference -- Multiclass}
\begin{algorithmic}[1]
\State \textbf{Input:} $\mathcal{R} = \{(x_i, y_i)\}_{i=1}^R$, $\mathcal{T} =  \left \{ \textbf{x}_p \right \}_{p=1}^{P}$, batch size $B$, class count $C$
\State Load model = $\phi_{\theta}$; $M_{\theta}$
\State Form $X = (\textbf{x}_1,\cdots, \textbf{x}_P)^T$ from $\mathcal{T}$

\State $F \gets \begin{bmatrix} \text{ } \end{bmatrix}$

\For{$k = 1$ to C}
    \State $\mathbf{y} \gets \begin{bmatrix} \text{ } \end{bmatrix}, S_k \gets \begin{bmatrix} \text{ } \end{bmatrix}$
    \State Create $\acute{\mathcal{R}}$ from $\mathcal{R}$ with new labels $y_i^k$ 
    \State Partition $\acute{\mathcal{R}}$ into $R = \lfloor |\acute{\mathcal{R}}| / B \rfloor$ batches $\{(X_r,\textbf{y}_r)\}_{r=1}^R$
    
    \State $Z = \phi(X,k)$
    \For{$r = 1$ to $R$}
        \State $Z_r = \phi(X_r,k)$
        \State Compute $\grave{S} = K_{(\theta;M)}(Z,Z_r)$
        \State $S_k \gets \begin{bmatrix} S_k & \grave{S} \end{bmatrix}$
        \State $\mathbf{y} \gets \begin{bmatrix} \mathbf{y} \\ \mathbf{y}_r \end{bmatrix}$
    \EndFor
    \State $\textbf{f}_k = S_k\textbf{y}\left (  \textbf{y}^{T}\textbf{y}\right )^{-1}$

    \State $F \gets \begin{bmatrix} F & \textbf{f}_k \end{bmatrix}$
\EndFor

\State $F_{i,:} = \operatorname{softmax}(F_{i,:})$

\State $\textbf{return}$  $\arg\max_{k} F_{:,k}$
\end{algorithmic}
\label{inference_algorithm_multiclass}
\end{algorithm}
}

\section{Experiments and Results}\label{sec:experiments}
\subsection{Datasets and Preprocessing}
Here, we use MIMIC-IV \cite{MoodyMIMIC-IVDatabase} dataset to empirically show the connection between CLaI-Binary and kernel methods. MIMIC-IV contains different types of patient data such as patient administrative data, physiological measurements, physician notes, and imaging data. For our experiments, we extract vital signs from patients diagnosed with sepsis to detect sepsis events. We follow the preprocessing procedure described in our previous work \cite{Haule2025CLaI:Prediction}, to extract multivariate recordings $X_i$ and its corresponding sample labels $\textbf{y}_i$. Note here that a recording is the same as a collection of samples and their respective labels as described in Section~\ref{sec:collaborative_learning}.

To evaluate multiclass classification we use three datasets: CIFAR10 \cite{abouelnaga_cifar-10_2016,krizhevsky_learning_2009, torralba_80_2008} (10 classes), PathMnist \cite{kather_predicting_2019} (9 classes), and the expanded SleepEDF dataset \cite{kemp_analysis_2000, goldberger_physiobank_2000} (5 classes). CIFAR10 consists of a large collection of 60000 tiny 32$\times$32 pixel colour images, to showcase the performance of CLaI-multiclass. The dataset, split into 50000 by 10000 train and test images respectively, comprises ten object categories, some of which are notably challenging to discriminate due to the limited information contained in low-resolution images \cite{abouelnaga_cifar-10_2016}. We separately normalise the training and testing datasets using mean and standard deviation of the dataset.

PathMNIST \cite{kather_predicting_2019} is one of the datasets in MedMNIST \cite{yang_medmnist_2021,yang_medmnist_2023}, a large collection of biomedical images at various resolutions for classification tasks. PathMNIST is a collection of about 107180 28$\times$28 images of slides stained with colorectal cancer tissue.

We use a subset of The expanded SleepEDF dataset\cite{kemp_analysis_2000, goldberger_physiobank_2000} containing 78 polysymnographic sleep recordings, using a bipolar electroencephalogram (EEG) montage, Fpz-Cz/Pz-Oz channels. We split the EEG recordings into 80/20 for training and testing. Each recording is segmented into 30 second non-overlapping windows along with their corresponding sleep stages: Wake (W), REM (R), non-Rem(N1,N2,N3,N4), movement (M) and ? (not scored). We merge N3 and N4 into N3 and drop M and ?, leading to a total 5 stages for multiclass classification \cite{deb_joint_2025}.



\subsection{Multiclass Classification}
\begin{figure*}[t]
\vspace{-3mm}
    \centering

    \begin{subfigure}{0.32\textwidth}
        \centering
        \includegraphics[width=0.85\linewidth]{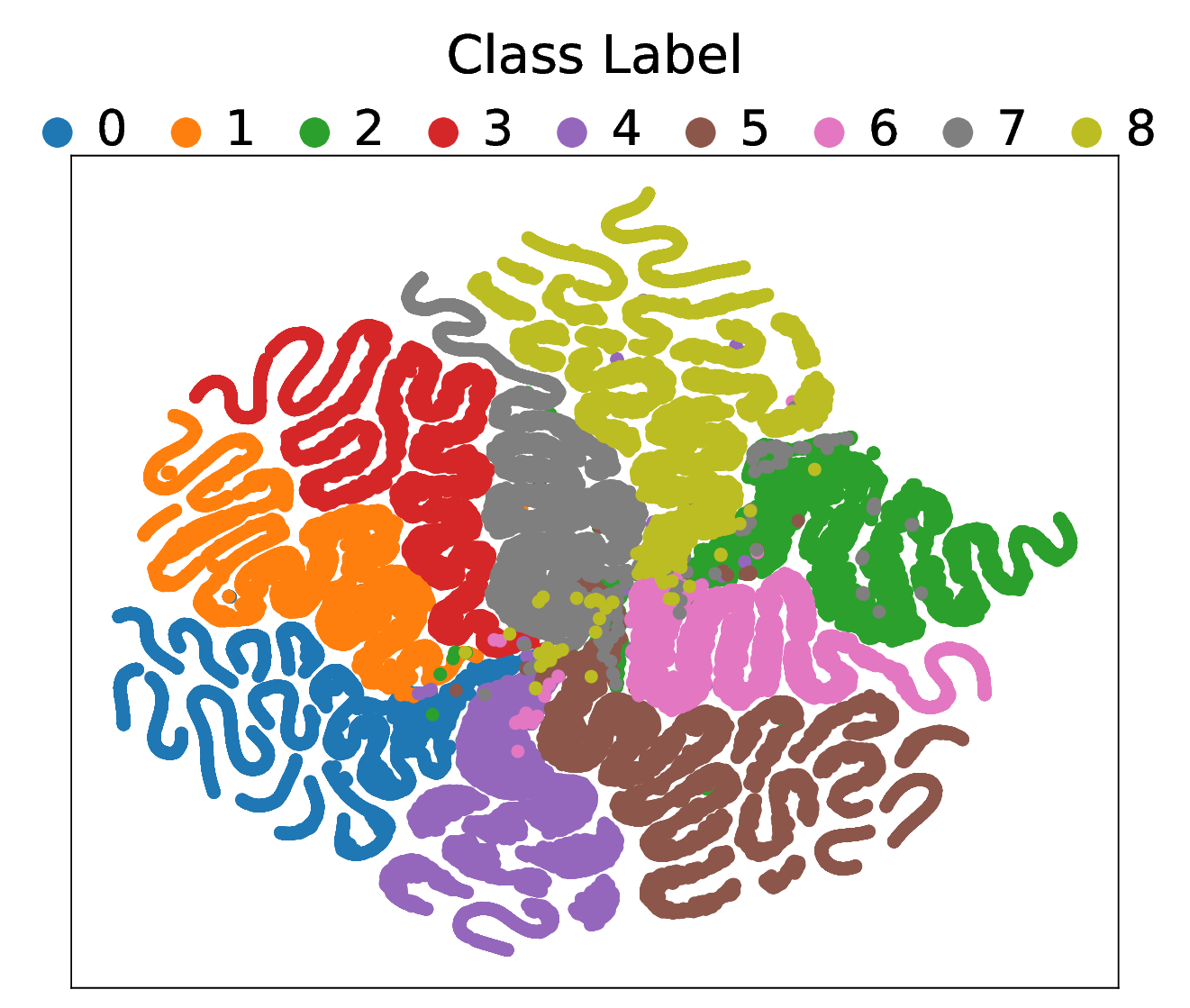}
        \caption{PathMNIST}
        \label{fig:pathmnist}
    \end{subfigure}
    \hfill
    \begin{subfigure}{0.32\textwidth}
        \centering
        \includegraphics[width=0.95\linewidth]{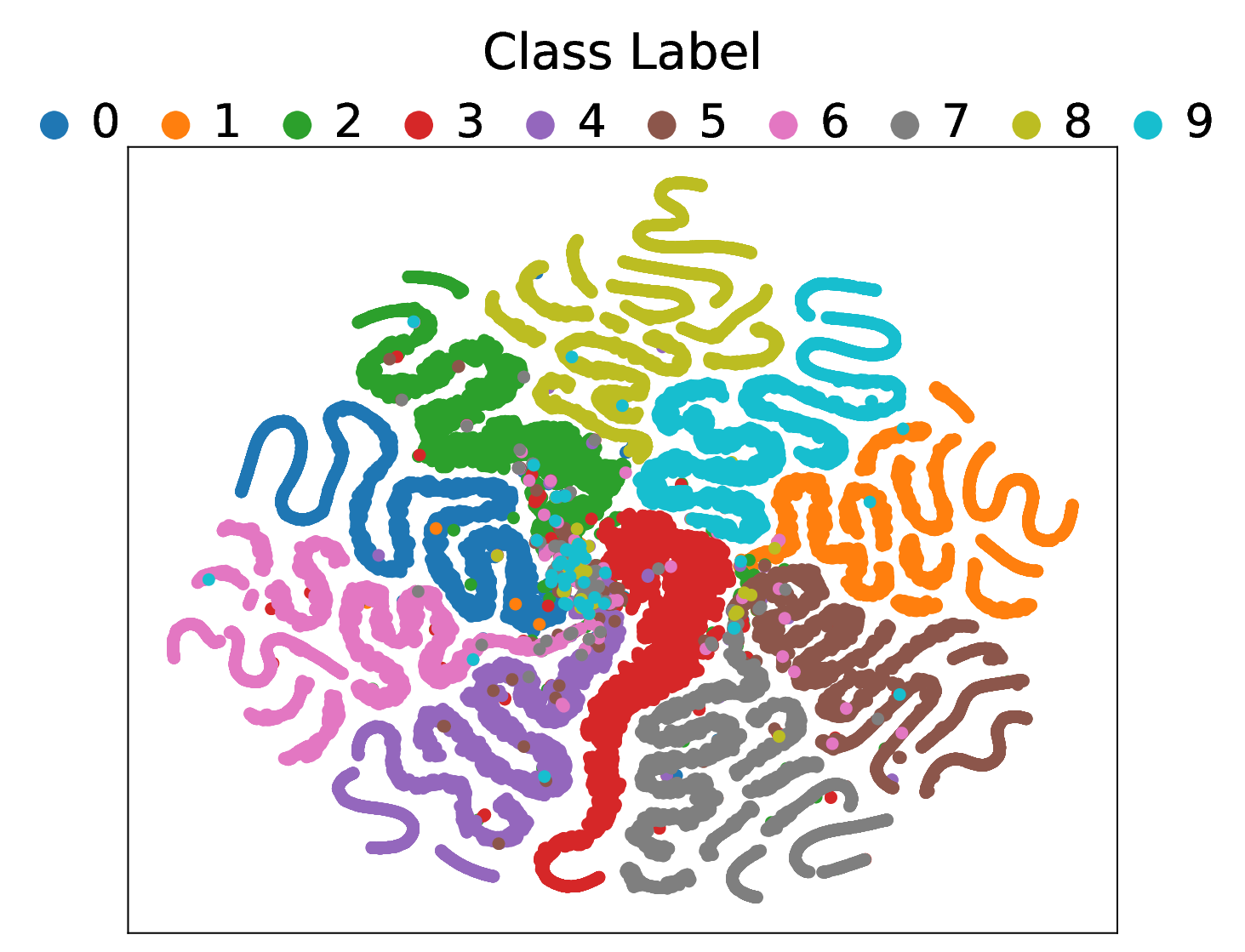}
        \caption{CIFAR-10}
        \label{fig:cifar10}
    \end{subfigure}
    \hfill
    \begin{subfigure}{0.29\textwidth}
        \centering
        \includegraphics[width=0.85\linewidth]{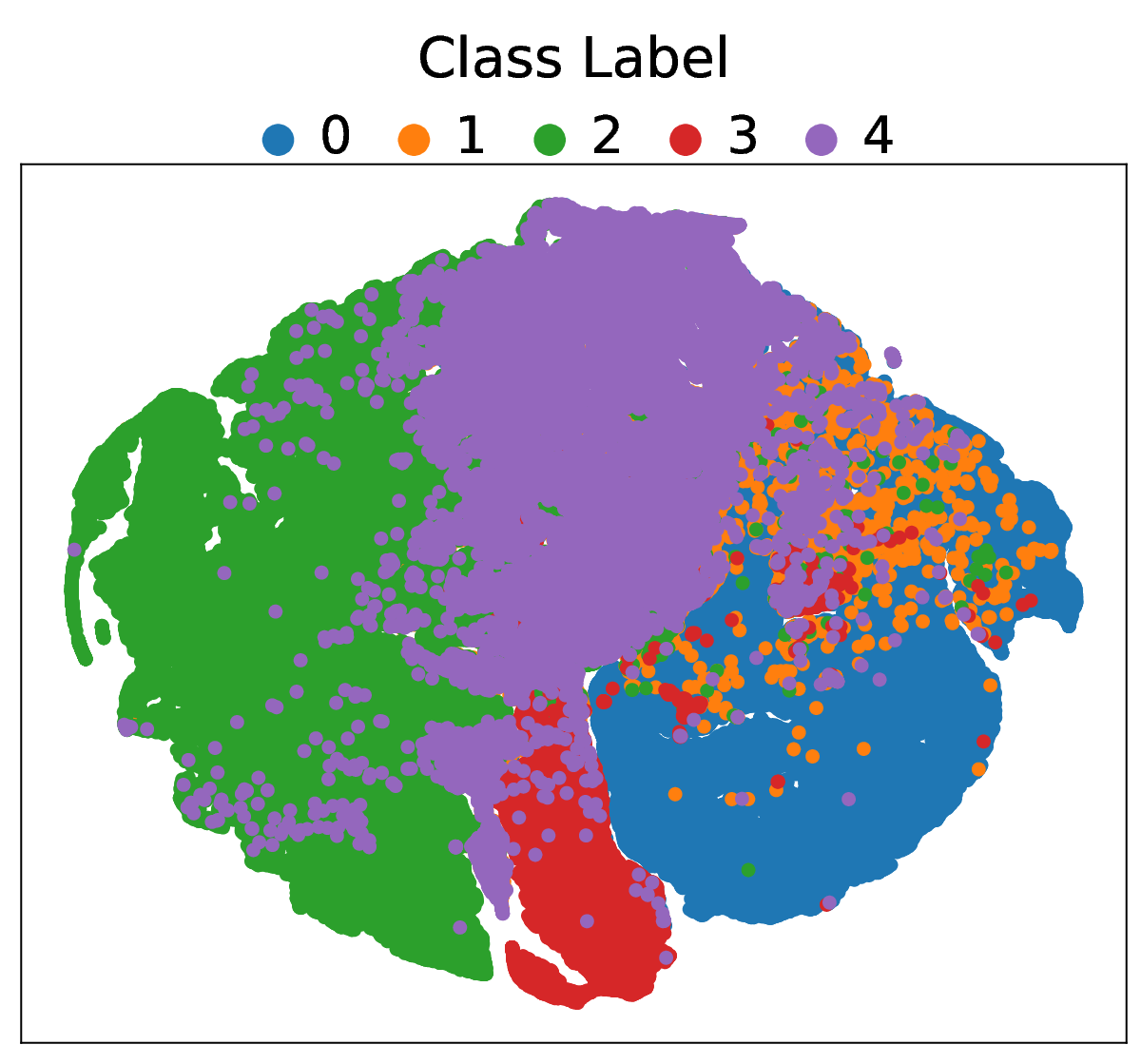}
        \caption{SleepEDF}
        \label{fig:sleepedf}
    \end{subfigure}

    \caption{Visualization of the latent space using t-SNE for the PathMNIST, CIFAR-10, and SleepEDF datasets.  Class index-to-label mappings are given in Table~\ref{tab:class_mappings}.}
    \label{fig:comparison}
\end{figure*}

We evaluate the multi-class extension of CLaI on image datasets, CIFAR10 \cite{krizhevsky_learning_2009} (10 classes), and PathMNIST \cite{kather_predicting_2019} (9 classes) using a lightweight ResNet-18-based embedding function, $\phi$, as a proof of concept. We note that a lightweight ResNet-18 model is not the state of the art in these computer vision tasks, but Table~\ref{tab:results} shows that our model achieves adequate performance compared to state-of-the-art methods and demonstrates the proof of principle of multiclas CLaI. Moreover, Fig.\ref{fig:comparison} illustrates that the latent space not only clusters samples of the same class but also groups similar classes together (e.g., four-legged animals: classes 5, 7, 3, 4; vehicles: classes 1, 9; blue background: classes 2, 0). Additionally, we evaluate CLaI on the SleepEDF dataset, with the results presented in Table~\ref{tab:results} and Fig.~\ref{fig:comparison}. The results indicate that CLaI achieves lower performance on SleepEDF than on the other evaluated datasets.

\begin{table}[t]
\centering
\caption{Classification performance on PathMNIST, CIFAR-10, and SleepEDF.}
\label{tab:results}
\small
\resizebox{0.9\columnwidth}{!}{%
\begin{tabular}{lccccc}
\toprule
Dataset   & Accuracy & Precision & Recall & F1-score & ROC-AUC \\
\midrule
PathMNIST & 79.97    & 81.74     & 79.97  & 79.75    & 92.82   \\
CIFAR-10  & 80.46    & 80.45     & 80.46  & 80.35    & 93.86   \\
SleepEDF  & 58.49    & 71.61     & 58.49  & 61.21    & 79.64   \\
\bottomrule
\end{tabular}
}
\end{table}

\begin{table*}[t]
    \centering
    \footnotesize
    \caption{Class index-to-label mappings and corresponding colors for the datasets shown in Figure~\ref{fig:comparison}.}
    \label{tab:class_mappings}
    \begin{tabular}{clc clc clc}
        \toprule
        \multicolumn{3}{c}{PathMNIST} & \multicolumn{3}{c}{CIFAR-10} & \multicolumn{3}{c}{SleepEDF} \\
        \cmidrule(r){1-3} \cmidrule(lr){4-6} \cmidrule(l){7-9}
        Idx & Label & Color & Idx & Label & Color & Idx & Label & Color \\
        \midrule
        0 & Adipose                          & \swatch{tab0} & 0 & Airplane   & \swatch{tab0} & 0 & Wake & \swatch{tab0} \\
        1 & Background                       & \swatch{tab1} & 1 & Automobile & \swatch{tab1} & 1 & N1   & \swatch{tab1} \\
        2 & Debris                           & \swatch{tab2} & 2 & Bird       & \swatch{tab2} & 2 & N2   & \swatch{tab2} \\
        3 & Lymphocytes                      & \swatch{tab3} & 3 & Cat        & \swatch{tab3} & 3 & N3   & \swatch{tab3} \\
        4 & Mucus                            & \swatch{tab4} & 4 & Deer       & \swatch{tab4} & 4 & REM  & \swatch{tab4} \\
        5 & Smooth Muscle                    & \swatch{tab5} & 5 & Dog        & \swatch{tab5} &   &      &               \\
        6 & Normal Colon Mucosa              & \swatch{tab6} & 6 & Frog       & \swatch{tab6} &   &      &               \\
        7 & Cancer-associated Stroma         & \swatch{tab7} & 7 & Horse      & \swatch{tab7} &   &      &               \\
        8 & Colorectal Adenocarcinoma Epith. & \swatch{tab8} & 8 & Ship       & \swatch{tab8} &   &      &               \\
          &                                  &               & 9 & Truck      & \swatch{tab9} &   &      &               \\
        \bottomrule
    \end{tabular}
\end{table*}


\subsection{Ablation of the learned similarity function}\label{sec:ablation_similarity_function}
\begin{figure}[t]
\vspace{-3mm}
    \centering
    \includegraphics[width=0.88\columnwidth]{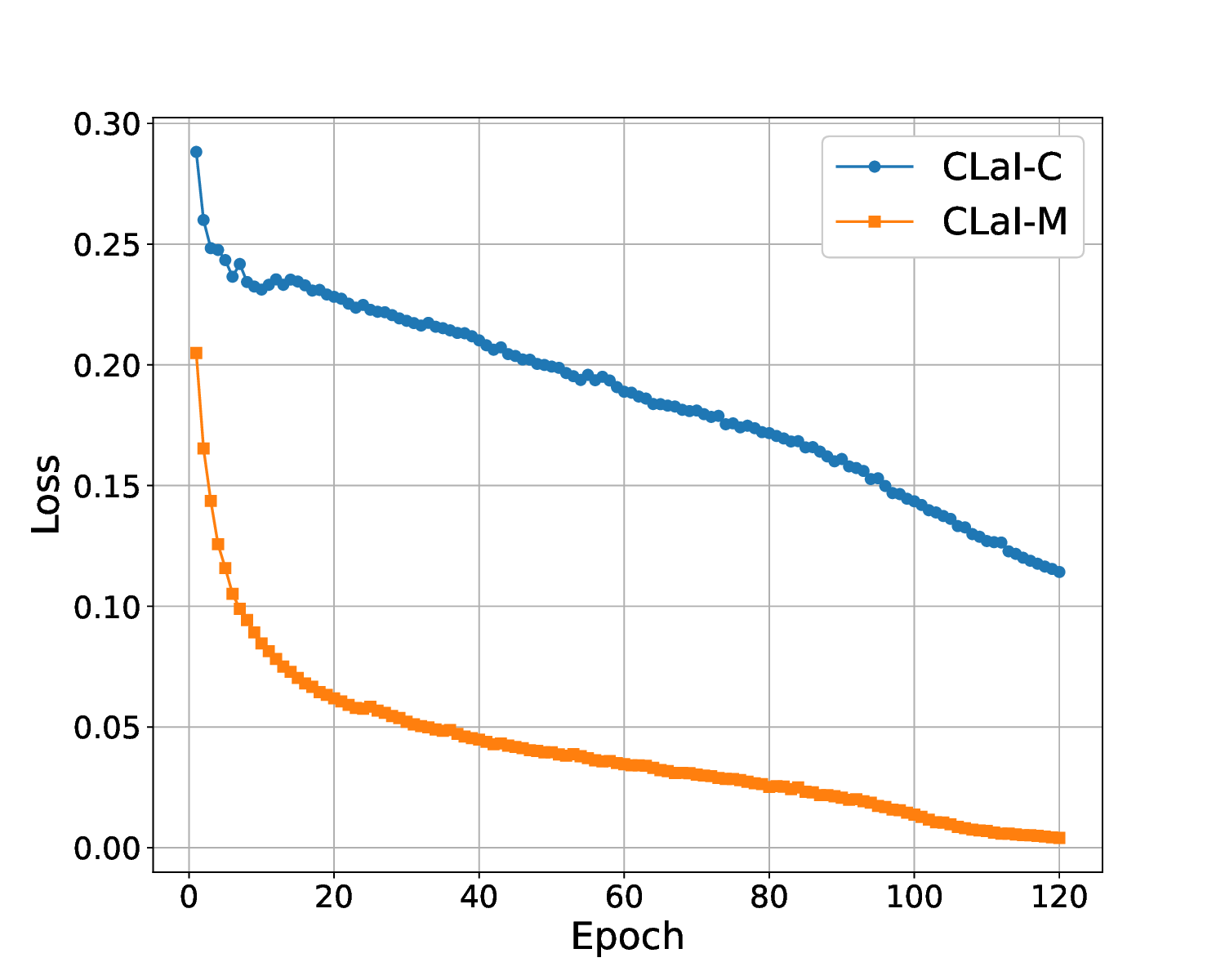}

    \caption{Training loss for CLaI with cosine similarity (CLaI-C) versus CLaI with Mahalanobis distance in the exponential base kernel (CLaI-M), showing that CLaI-M achieves a much smaller loss by epoch 120 than CLaI-C, i.e., faster convergence.}
    \label{fig:claiC_vs_claiM}
\end{figure}

Here we explore how the choice and flexibility of the similarity function affect CLaI. We compare cosine similarity kernel (CLaI-C) \eqref{eq:cosine_similarity}, which was used in our previous implementation of CLaI to Mahalanobis distance based kernel (CLaI-M), \eqref{eqn:mahalanobis}. Fig.~\ref{fig:claiC_vs_claiM}, shows that CLaI-M achieves a much lower loss by epoch 120 compared to CLaI-C, hence faster convergence.

\subsection{Connection to kernel methods}
In this section, we perform experiments to assess the quality of CLaI predictions in comparison to Kernel methods. Specifically, we visually explore the latent representation with respect to occurrence of events and calibration of predictions.
\subsubsection{Latent signals: CLaI vs. Gaussian Process}
\begin{figure*}[!t] \centering \begin{subfigure}[b]{0.49\textwidth} \centering \includegraphics[width=\linewidth]{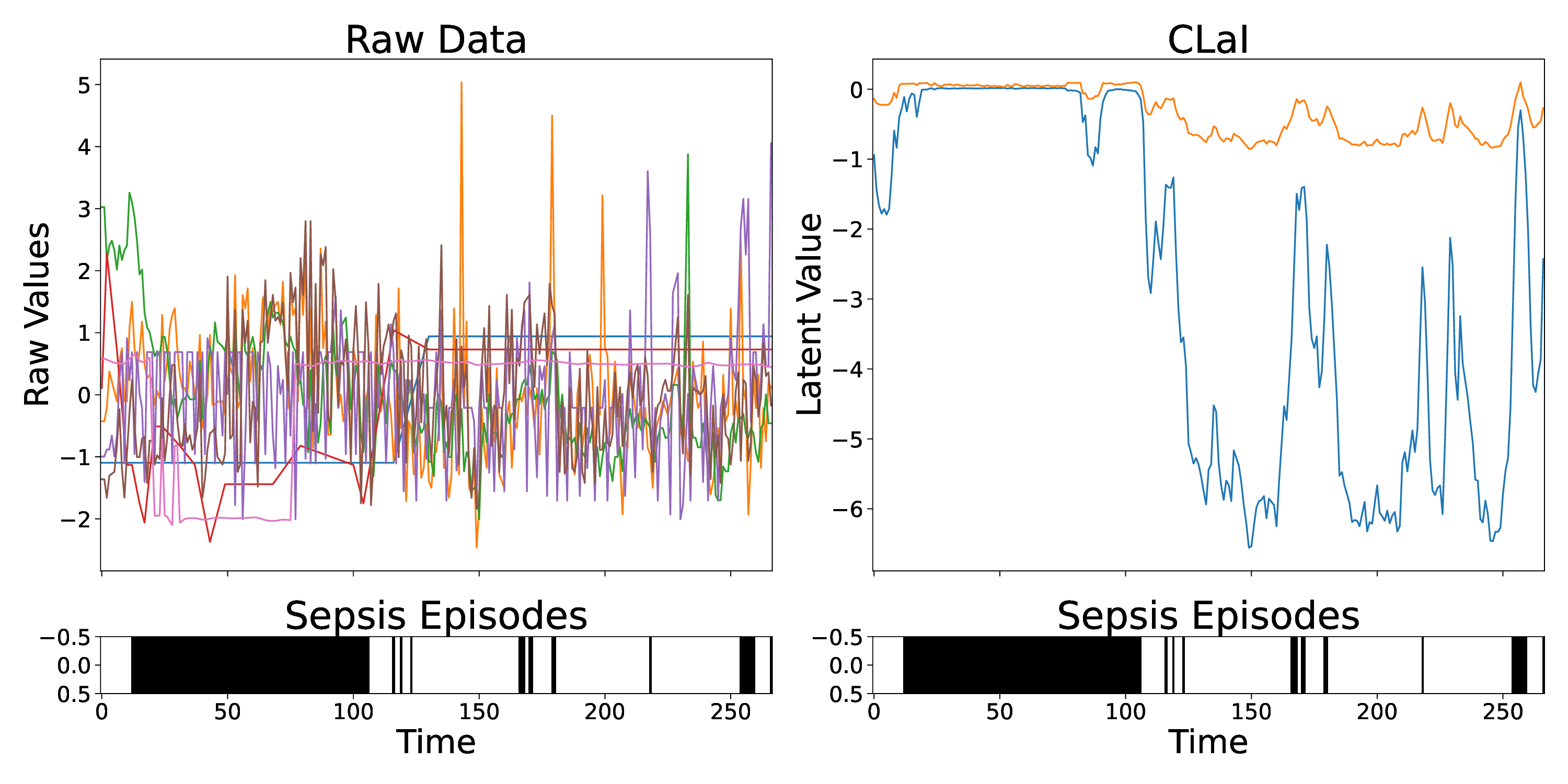} \caption{} \label{fig:latent_signals_clai_main} \end{subfigure} \hfill \begin{subfigure}[b]{0.49\textwidth} \centering \includegraphics[width=\linewidth]{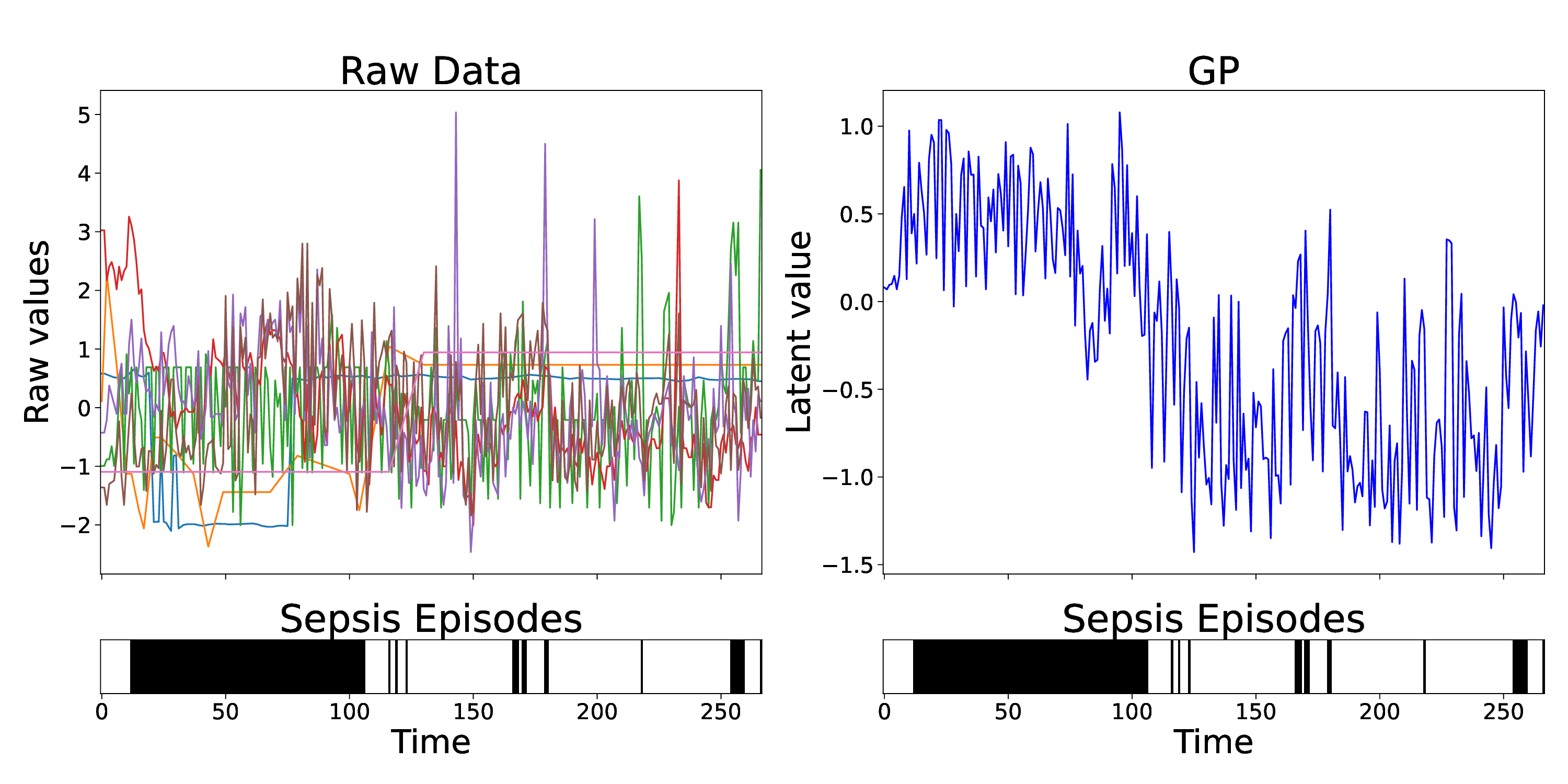} \caption{} \label{fig:latent_signals_gp_main} \end{subfigure} \caption{Comparison of latent signals generated by CLaI and GP for the same patient stay. The CLaI latent signals converge during events, resulting in smaller areas between them, and diverge when no event is present, increasing the area between the signals. In contrast, the GP latent signal increases during an event but contains more noise than the signals generated by CLaI.} \label{fig:latent_signals_comparison} \end{figure*}

Although CLaI and GP classification arise from different objectives, both construct predictions through similarity-weighted contributions from reference/training samples. We therefore compare their latent signals.
Here, we plot the embedding signals to examine their behaviour upon event occurrence. Fig.~\ref{fig:latent_signals_clai_main} shows that latent signals converge during sepsis events, resulting in smaller areas between them. In contrast, they diverge when no event is present, increasing the area between the signals. For comparison, we visualise the latent signal generated by a Gaussian Process model in Fig.~\ref{fig:latent_signals_gp_main}, showing that the latent signal value increases during an event. However, the figure shows that the signal is not as smooth as that generated by CLaI.

\subsubsection{Calibration of CLaI in sepsis prediction}
We evaluate the calibration of the CLaI framework on sepsis prediction by comparing two CLaI variants that differ in their similarity measures: one using a cosine similarity kernel and another employing an exponential kernel based on the Mahalanobis distance, as described in section~\ref{sec:mahalanobis_kernel}. To assess the probabilistic fidelity of the predictions, we benchmark these results against a traditional Bayesian approach using a mixture model of Student‑$t$ distributions, as we found that the embeddings resulting from collaborative learning tend to exhibit long-tailed distributions. Calibration is quantified using Expected Calibration Error (ECE), which is a widely used metric that measures how well a model's prediction confidence matches the observed frequency of correctness. In addition we compute other performance metrics including AUROC, AUPRC and F1 scores, and report the results in Table~\ref{tab:detection_UQ_experiments}.

\begin{table}[!ht]
\caption{Sepsis detection performance results across different experiments involving different versions of CLaI and the benchmark model MM of Student-$t$. The results show that CLaI with Mahalanobis distance kernel (CLaI-M) has lowest ECE, despite having a slightly lower performance across other metrics. Moreover CLaI achieves the highest performance and calibration in comparison to the mixture model (MM). ECE quantifies the calibration of the model's predicted probabilities.}
\label{tab:detection_UQ_experiments}
\resizebox{\columnwidth}{!}{%

\begin{tabular}{lcccc}
\toprule
\textbf{Experiment} & \textbf{AUROC} & \textbf{AUPRC} & \textbf{F1} & \textbf{ECE} \\
\midrule
\midrule
\multirow{1}{*}{CLaI-M} 
    & 0.804 (0.014) & 0.635 (0.035) & 0.613 (0.055) & \textbf{0.229 (0.029)} \\
\midrule
\multirow{1}{*}{CLaI-C} 
    & \textbf{0.848 (0.009)} & \textbf{0.714 (0.019)} & \textbf{0.701 (0.011)} & 0.262 (0.006) \\
\midrule
\multirow{1}{*}{MM of Student-$t$}
    & 0.835 (0.016) & 0.701 (0.027) & 0.699 (0.026) & 0.278 (0.023) \\
\bottomrule
\end{tabular}
}

\end{table}

\section{Discussion}\label{sec:discussion}

This work presents a general formulation of Collaborative Learning and Inference (CLaI) that establishes connections between kernel alignment, learned representations, and kernel-based classification. The framework can be viewed as a two-stage process: Collaborative Learning learns an embedding whose induced similarity function is aligned with a label-derived target kernel, while Collaborative Inference uses this learned similarity structure to predict the labels of unseen samples. We evaluated this formulation across binary \cite{Haule2025CLaI:Prediction} and extended to multiclass classification tasks, including PathMNIST, CIFAR-10, SleepEDF, and sepsis prediction.

A central result of this work is the interpretation of Collaborative Learning as a kernel alignment procedure. Rather than prescribing a similarity function independently of the task, the embedding model is trained so that pairwise similarities in the learned representation agree with the desired class-based similarity structure. The resulting embedding therefore induces a kernel that is explicitly shaped by the classification objective. This provides a direct connection between representation learning and kernel alignment, and offers an alternative to selecting a kernel a priori.
This interpretation extends naturally to Collaborative Inference. The inference rule can be expressed as a similarity-weighted combination of the labels of a reference set. As shown in Section~\ref{sec:collaborative_inference_bayes}, this has the same form as a kernel Bayes classifier when the class-conditional densities are estimated using Parzen window methods. The empirical proportions of the reference classes act as estimates of the class priors, while the sums of kernel similarities within each class provide estimates of the corresponding class-conditional densities. Collaborative Inference can therefore be interpreted as kernel Bayes classification operating in a learned representation space.

The multiclass experiments demonstrate that this formulation extends beyond binary classification. Across PathMNIST, CIFAR-10, and SleepEDF, CLaI was able to learn representations that supported effective multiclass prediction. The learned embeddings also exhibited meaningful class structure, with samples from the same class tending to cluster while visually or semantically related classes remained closer in the latent space. We see these class groups in Fig.~\ref{fig:cifar10}, vehicles(truck and automobile), four-legged animals (dog, horse, cat, dear), blue-background (ship, bird, airplane). Note that the frog class sits between blue-background and four-legged animals groups. This suggests that the alignment objective does not simply separate classes independently, but can also capture relationships between classes through the learned similarity structure.

The choice of similarity function was found to have an important effect on the behaviour of CLaI. Replacing cosine similarity with a Gaussian kernel based on a learned Mahalanobis distance resulted in improved multiclass performance and faster convergence in our experiments. The Mahalanobis formulation also produced lower calibration error than the cosine-based formulation in the sepsis prediction experiment. These results suggest that allowing the geometry of the latent space to be learned jointly with the embedding can provide a more expressive similarity function than cosine similarity alone. In particular, the Gaussian kernel induces an infinite-dimensional feature space, and hence a richer RKHS, whereas cosine similarity applied to a finite-dimensional embedding corresponds to a finite-dimensional kernel representation \cite{scholkopf_learning_2001}. Importantly, this extension retains the kernel-alignment formulation while providing additional flexibility in the structure of the learned similarity.

The relationship between CLaI and established kernel methods was further explored through the auxiliary experiments. Although CLaI and Gaussian-process classification \cite{Rasmussen2005GaussianLearning} are derived from different optimisation objectives, their predictive structures share an important commonality: both produce predictions from similarity-weighted contributions of reference samples. The observed similarity between their latent signals is consistent with interpreting CLaI within the broader family of kernel-based learning methods, and complements the theoretical result that Collaborative Inference implements a kernel Bayes classifier in the learned representation space.

The differences between the approaches, however, are equally important. Both Gaussian-process classification and DKL \cite{wilson_deep_2016} optimise kernel parameters by maximising the marginal likelihood of a probabilistic model over functions. CLaI departs from this family in its objective: rather than maximising a likelihood, it aligns the induced kernel directly with a label-derived target matrix, optimising the learned similarity for discriminative class structure. CLaI is closest to DKL in that both learn a feature representation that induces the kernel rather than fixing it in advance; the distinction lies in the objective, i.e. alignment \cite{cristianini_kernel-target_2001} versus marginal likelihood. Consequently, CLaI explicitly shapes the learned similarity for discriminative class structure, whereas GP- and DKL-based methods are formulated within a probabilistic Gaussian-process framework. This positions CLaI as a complementary member of the learned-kernel family, combining representation learning with a direct kernel-alignment objective.

Overall, these results support a unified interpretation of CLaI as a learned-kernel framework: Collaborative Learning learns a task-specific similarity function through kernel alignment, and Collaborative Inference subsequently performs kernel-based classification using that learned similarity. This provides a common perspective on the training and inference stages and connects the framework to established ideas from kernel methods, Bayesian classification, and deep kernel learning.

\section{Limitations}\label{sec:limitations}

The principal limitation of CLaI is its computational complexity, which is more pronounced in the multiclass extension. Because Collaborative Learning compares every batch against every other batch, its time complexity is quadratic in the number of batches, $\mathcal{O}(T\cdot M^2)$, where $T$ and $M$ denote the number of epochs and batches, respectively. The multiclass extension introduces an additional loop that conditions the embedding model on each class in turn, raising the cost to 
$\mathcal{O}(T\cdot C\cdot M^2)$, where $C$ is the number of classes. (For clarity, these expressions omit the per-pair cost.) Storage complexity, by contrast, is unaffected by the number of classes, since the class loop executes sequentially and reuses the same buffers; it remains $\mathcal{O}(P + MA + B^2 +BL)$, where $P$, $A$, $B$, and $L$ denote the number of model parameters, the activations stored per batch for one backward pass, the batch size, and the latent dimension, respectively (ignoring kernel-specific costs). Owing to these costs, we deliberately used lightweight embedding models as a proof of concept for multiclass classification, rather than aiming to surpass state-of-the-art performance.

\section{Conclusion and Future Work}\label{sec:conclusion}
In this work we showed that CLaI is, in principle, a kernel alignment method, and we extended it to the multiclass setting. Our experiments verify these theoretical claims and establish a proof of concept for multiclass classification.

A natural direction for future work is to reduce the computational cost of CLaI, particularly for multiclass applications. The additional loop over the $C$ classes, which conditions the embedding model $\phi$ on each class in turn, is the main source of this overhead. This loop could be eliminated by computing a single target/similarity tensor of shape $(C, B, B)$, replacing the
$C$ sequential passes with a single pass, possibly supported by an additional preprocessing step to prepare the tensors.

Having established CLaI as an instance of supervised kernel learning by alignment, another promising direction is to extend the framework to unsupervised kernel learning by alignment \cite{gao_kernel_2021,insulla_towards_2024,sui_unicon_2025,lin_unsupervised_2024,sun_unsupervised_2025}. Combined with the DKL capabilities that CLaI already offers, such an extension would be especially valuable in domains where labelled data are scarce, such as medical applications.

Finally, we plan to conduct further experiments exploring interpretability through case-based reasoning, with the aim of demonstrating CLaI as a practical and adoptable solution for advancing healthcare delivery.



\section*{Acknowledgement}\label{sec:ack}
The authors are thankful to Dr Alfredo Gonzalez-Sulser for his revision of this manuscript.


\bibliographystyle{ieeetr}
\bibliography{references_zotero,references_mendeley}
\end{document}